\documentclass[conference]{IEEEtran}
\IEEEoverridecommandlockouts
\usepackage{cite}
\usepackage{amsfonts,amssymb}
\usepackage{amsmath}
\usepackage{booktabs}
\usepackage{multirow}
\usepackage{CJKutf8}
\usepackage{amsmath,amssymb,amsfonts}
\usepackage{algorithmic}
\usepackage{graphicx}
\usepackage{textcomp}
\usepackage{xcolor}
\def\BibTeX{{\rm B\kern-.05em{\sc i\kern-.025em b}\kern-.08em
    T\kern-.1667em\lower.7ex\hbox{E}\kern-.125emX}}
\begin{document}

\title{Ensuring Consistency for In-Image Translation}

\author {
    \IEEEauthorblockN{
    Chengpeng Fu\IEEEauthorrefmark{1}\IEEEauthorrefmark{2},
    Xiaocheng Feng\IEEEauthorrefmark{1}\IEEEauthorrefmark{2},
    Yichong Huang\IEEEauthorrefmark{1},
    Wenshuai Huo\IEEEauthorrefmark{1}\IEEEauthorrefmark{2},
    Baohang Li\IEEEauthorrefmark{1},
    Zhirui Zhang,
    Yunfei Lu,\\
    Dandan Tu,
    Duyu Tang,
    Hui Wang\IEEEauthorrefmark{2},
    Bing Qin\IEEEauthorrefmark{1}\IEEEauthorrefmark{2}, 
    Ting Liu\IEEEauthorrefmark{1}\IEEEauthorrefmark{2}
    }
    \IEEEauthorblockA{\IEEEauthorrefmark{1} Harbin Institute of Technology, Harbin, China. \{cpfu, xcfeng, ychuang, wshuo, qinb, tliu\}@ir.hit.edu.cn}
    \IEEEauthorblockA{\IEEEauthorrefmark{2} Pengcheng Laboratory, Shenzhen, China. wangh06@pcl.ac.cn}
}

\maketitle

\begin{abstract}
The in-image machine translation task involves translating text embedded within images, with the translated results presented in image format. While this task has numerous applications in various scenarios such as film poster translation and everyday scene image translation, existing methods frequently neglect the aspect of consistency throughout this process. We propose the need to uphold two types of consistency in this task: translation consistency and image generation consistency. The former entails incorporating image information during translation, while the latter involves maintaining consistency between the style of the text-image and the original image, ensuring background integrity. To address these consistency requirements, we introduce a novel two-stage framework named \textbf{HCIIT} (\textbf{H}igh-\textbf{C}onsistency \textbf{I}n-\textbf{I}mage \textbf{T}ranslation) which involves text-image translation using a multimodal multilingual large language model in the first stage and image backfilling with a diffusion model in the second stage. Chain of thought learning is utilized in the first stage to enhance the model's ability to leverage image information during translation. Subsequently, a diffusion model trained for style-consistent text-image generation ensures uniformity in text style within images and preserves background details. A dataset comprising 400,000 style-consistent pseudo text-image pairs is curated for model training. Results obtained on both curated test sets and authentic image test sets validate the effectiveness of our framework in ensuring consistency and producing high-quality translated images.
\end{abstract}


\section{Introduction}
\label{sec:intro}

With globalization on the rise, the demand for machine translation in our lives is increasing significantly. Beyond conventional text translation, we frequently encounter instances where text embedded within images requires translation, spanning from movie posters and children's picture books to various everyday situations. Currently, research focused on translating text within images, such as Text Image Translation (TIT) \cite{ma2022improving,lan2023exploring}, aims to precisely convert text from source images into the desired target text, rather than integrating the translated text back onto the source images. However, in practical scenarios, there is an increasing demand to display the translated text in image form to offer a more intuitive representation. We refer to this task of rendering text back onto images as In-Image Translation (IIT) in this work. 

Commercial systems such as Google and Aliyun typically present results in image format, focusing primarily on translations for everyday practical scenarios. In these scenarios, the emphasis lies on conveying the placement and significance of the translated text within the image, rather than prioritizing the visual smoothness of the rendered image. \cite{qian-etal-2024-anytrans} have proposed the AnyTrans framework to improve the fluency of translated images. However, all these approaches overlook the consistency assurance in the process of IIT. We emphasize that in conducting IIT, the following consistencies should be ensured to meet the requirements of most scenarios:
\begin{itemize}
\item Translation Consistency: Image information should be integrated into the translation process. While previous studies on multimodal machine translation \cite{caglayan2016multimodal,yao2020multimodal,yin2020novel,caglayan2021cross,li2022vision} have primarily focused on enhancing text translation quality by leveraging image information during the text translation process, text translation accompanied by images is relatively uncommon in real-life scenarios. In IIT, text is always accompanied by image, making image information particularly crucial during the translation process.
\item Image Generation Consistency: In certain scenarios such as film poster translation and children's picture book translation, we aim for the style of the target text to be consistent with that of the source text. This consistency includes aspects like font and color consistency in the text images, as well as minimal alteration to the background. Existing methods like Anytrans determine the style of the generated text image based on the surrounding context of the rendered image. A more appropriate approach would be to ensure consistency with the style of the input source text image.
\end{itemize}

\begin{figure}[t] 
\centering 
\includegraphics[width=0.5\textwidth]{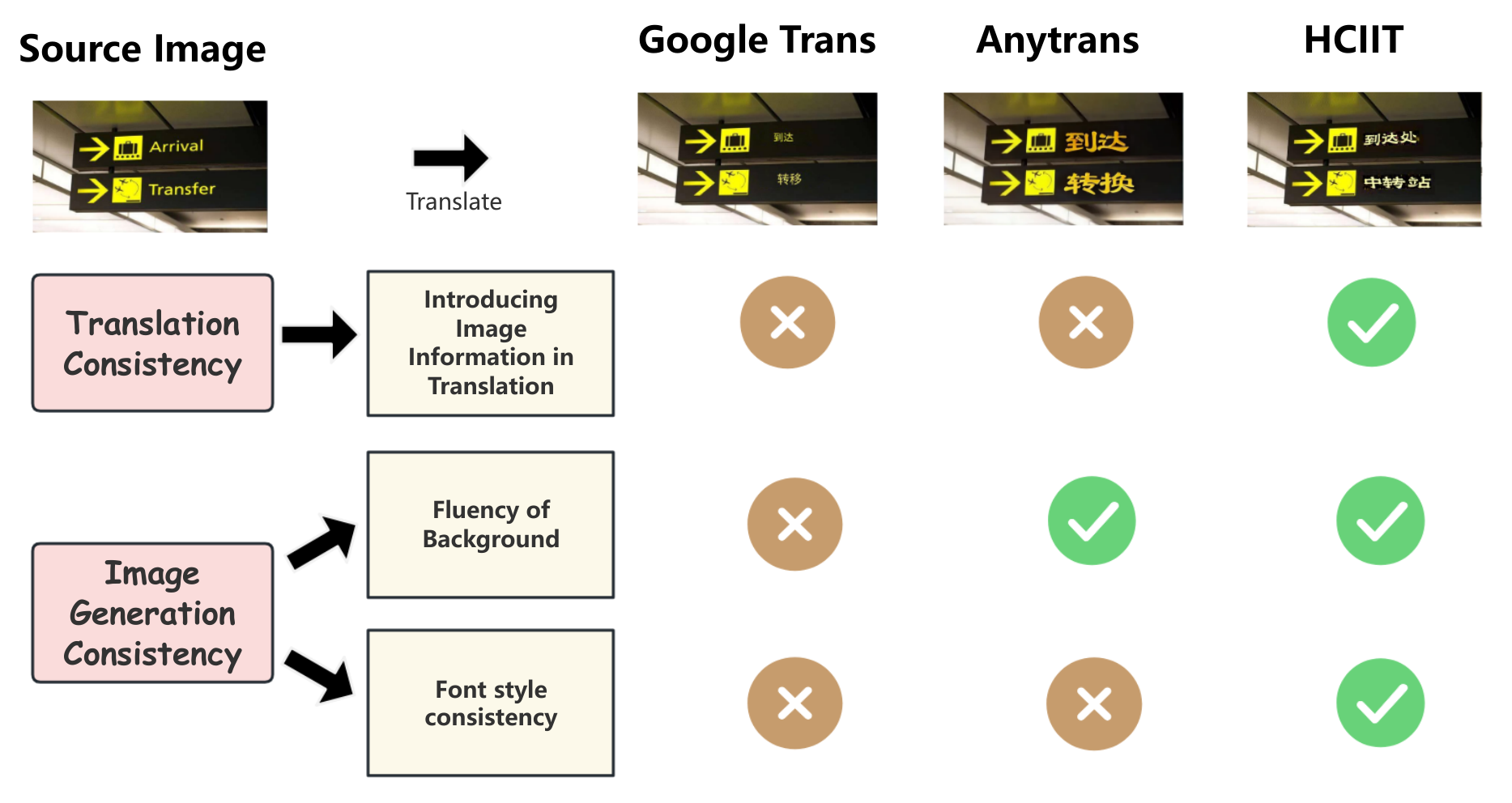} 
\caption{Comparison of the performance of our method, Anytrans and online systems in terms of translation consistency and image generation consistency. "Transfer" in the image should be translated as \begin{CJK*}{UTF8}{gbsn} "中转" \end{CJK*} (transfer station/transit) not \begin{CJK*}{UTF8}{gbsn} "转移/转换" \end{CJK*} (convert/shift). The style of the target image should be consistent with the source image. More examples can be find in Figure \ref{fig:5}.} 
\label{fig:1} 
\end{figure}

As shown in Figure \ref{fig:1}, the existing methods are unable to fully meet these consistency rules, so we further propose a two-stage framework for IIT called HCIIT: (1) TIT based on a Multimodal Multilingual Large Language Model(MMLLM): In this stage, we leverage Chain of Thought (CoT) learning to ensure effective utilization of image information during the translation process, thereby enhancing translation accuracy. (2) Translation Result Backfilling using Diffusion Model: In this stage, we use a diffusion model to generate images, which can produce high-quality images, ensuring background smoothness. However, existing diffusion models still have certain issues in text-image generation and struggle to generate stylistically consistent text images. Therefore, we introduce a style latent module to the existing text-control diffusion model to enhance the quality of text images and learn styles effectively. Additionally, since real translation image pairs are scarce in real-life settings, we generate synthetic data with different styles (including various fonts, text sizes, colors, and italicization) for style consistency learning. By evaluating our method on both synthetic test sets and real image test sets, we observed superior translation performance and improved style consistency compared to previous commercial systems and methods. In essence, our work presents two key contributions: 
\begin{itemize}
\item We underline current inadequacies in maintaining consistency in in-image translation tasks, with a specific emphasis on ensuring consistency in translation and image generation. We also propose a framework to enhance these consistencies.
\item We have curated over 400,000 style-consistent pseudo-parallel image pairs and trained a model capable of generating image-with-text content in a consistent style. Our framework has yielded satisfactory and acceptable results in ensuring consistent.
\end{itemize}






\begin{figure*}[h] 
\centering 
\includegraphics[width=0.9\textwidth]{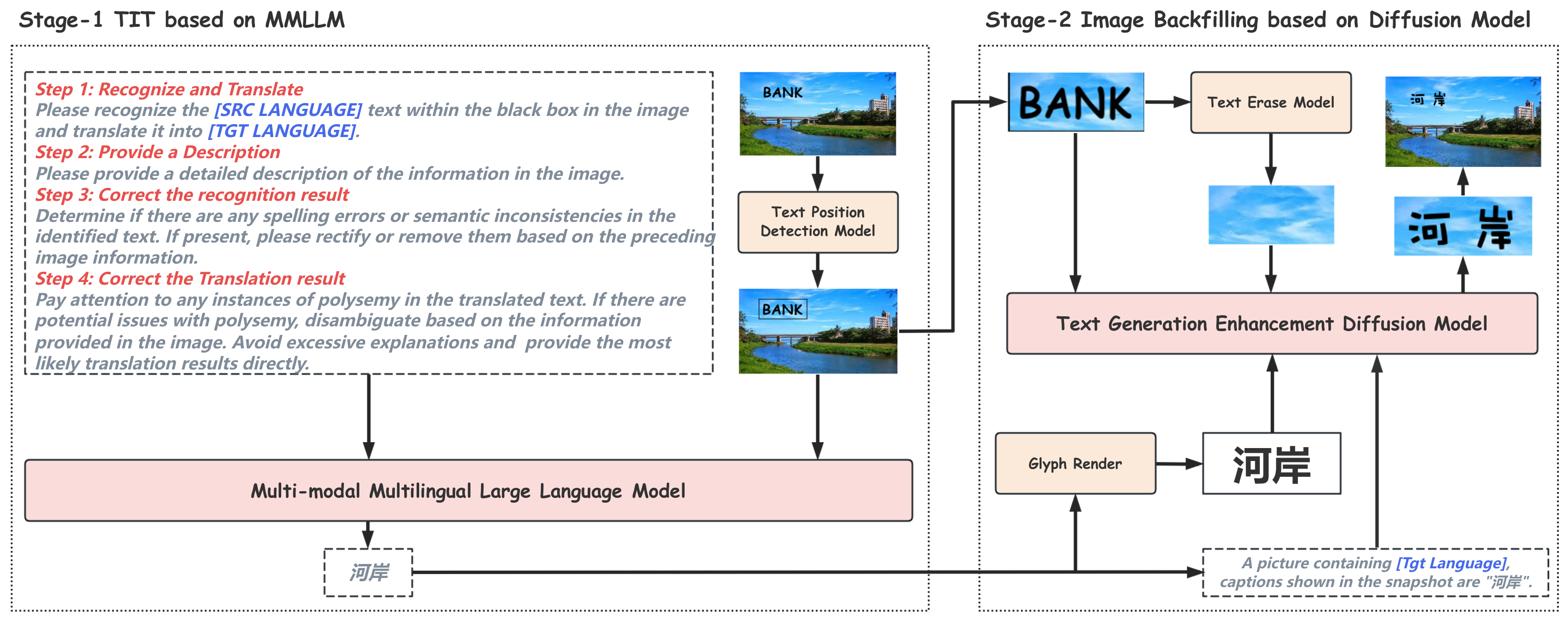} 
\caption{The process of two-stage in-image translation with consistent style. Our framework consists of two stages, comprising an MMLLM-based TIT and a diffusion-model-based image backfilling.} 
\label{fig:2} 
\end{figure*}

\section{Related Work}
\subsection{In-Image Translation}
Text Image Translation (TIT) \cite{watanabe1998translation,chen2015integrating,ma2022improving,lan2023exploring}, which is a subtask of IIT, aims to translate text on images into the target language, primarily returning the translation in textual form. For returning translation results in image form, AnyTrans proposed utilizing a MMLLM for translation, and then employing Anytext to render the text onto images. However, it did not focus on integrating image information during translation, and it also did not maintain consistent text styles. Our framework addresses these limitations. Current methods for IIT still rely on cascaded approaches, while some have proposed a method like Translatotron-V \cite{lan2024translatotron} to achieve end-to-end translation, primarily for text images with easy and solid backgrounds. A crucial future challenge is how to achieve high-quality end-to-end translation for complex text images in natural scenes with intricate backgrounds.
\subsection{Text Image Generation}
In our framework, we utilize text-to-image diffusion models to generate target images. However, existing methods often struggle to produce high-quality text images. Several efforts have been made to enhance the text generation capabilities of diffusion models: GlyphDraw \cite{ma2023glyphdraw} introduces glyph images as conditions to improve text image generation. GlyphControl 
 \cite{yang2024glyphcontrol} further incorporates text position and size information. TextDiffuser \cite{chen2024textdiffuser} constructs a layout network to extract character layout information from the text prompt, generating a character-level segmentation mask as a control condition. TextDiffuser-2 \cite{chen2023textdiffuser} builds upon TextDiffuser by unleashing the power of language models for text rendering. Anytext \cite{tuoanytext} transforms the text encoder through the integration of semantic and glyph information. Additionally, it incorporates an OCR encoder to supervise text generation at a more granular level. All of these methods mentioned so far struggle to generate stylistically consistent images. Building upon Anytext, we further modified the framework by incorporating a style module to ensure the consistency of text styles in the generated images.

\section{Methodology}
In this section, we will initially present our two-stage framework, designed to ensure consistency in IIT. Subsequently, we will provide detailed explanations of the methods employed in each of the two stages.

\subsection{High-Consistent In-Image Translation Framework}
We propose a two-stage methodology to ensure high consistency in IIT, as depicted in Figure \ref{fig:2}: (1) Text image translation based on a MMLLM. For a source image $S$, the initial step involves text position recognition, which yields $N$ pieces of information regarding the locations of text elements. These details are presented in the form of text boxes and denoted as $P=\{p_i\}^N_{i=1}$. Next, we will individually feed source image $S$ and the position of each text element $p_i$ into a MMLLM to accomplish text recognition and translation, resulting in the translated output $Y=\{y_i\}^N_{i=1}$. This large language model must support the source and target languages involved in our translation process. (2) Image backfilling using a stable diffusion model. In this stage, we utilize a diffusion model to generate an image $T$ containing the translated text $Y$ while maintaining the style consistent with $S$. We transition the outputs from the first stage to the four inputs of the second stage: 1) For each translated text $y_i$ in $Y$, we create an image $G_i$ rendered in a standard font. 2) Simultaneously, based on $y_i$, we generate a prompt $y^p_i$ through rules to guide the diffusion model in generating the corresponding image. 3) Using the positional information $P_i$, we extract text from the image to obtain image 
bounding box $S^p_i$. 4) Additionally, we remove the text in $S^p_i$ to derive the background image $S^b_i$, and then utilize this information to generate images $T^p_i$ corresponding to $y_i$ and $S^p_i$. Finally, we fill in these $T^p_i$ using position $P_i$ to obtain the ultimate translated target image $T$.

\subsection{Stage 1: TIT based on MMLLM.} The first stage focuses on text recognition and translation on images using a MMLLM. We employ this model to accomplish the task, leveraging its advanced image understanding capabilities. However, we observed that when applied to image translation, these models tend to produce literal translations. For instance, when confronted with the word "Bank" in this image in Figure 2, a direct translation would yield \begin{CJK*}{UTF8}{gbsn} "银行" \end{CJK*}(financial institution). However, a more accurate translation in the context of the image would be \begin{CJK*}{UTF8}{gbsn} "河岸" \end{CJK*}(riverbank). Therefore, while MMLLMs excel at image comprehension, they struggle to effectively incorporate image information for accurate translation. \cite{huang2024aligning} also noted the misalignment between general comprehension abilities and translation comprehension abilities of large language models. To address this limitation in our framework, we adopt a CoT learning approach which forces the model to generate the most probable translation results based on its understanding of the image. The specific prompts we employed are illustrated in figure \ref{fig:2}. 

\subsection{Stage 2:Image Backfilling based on Diffusion Model.} 
\begin{figure*}[h] 
\centering 
\includegraphics[width=0.7\textwidth]{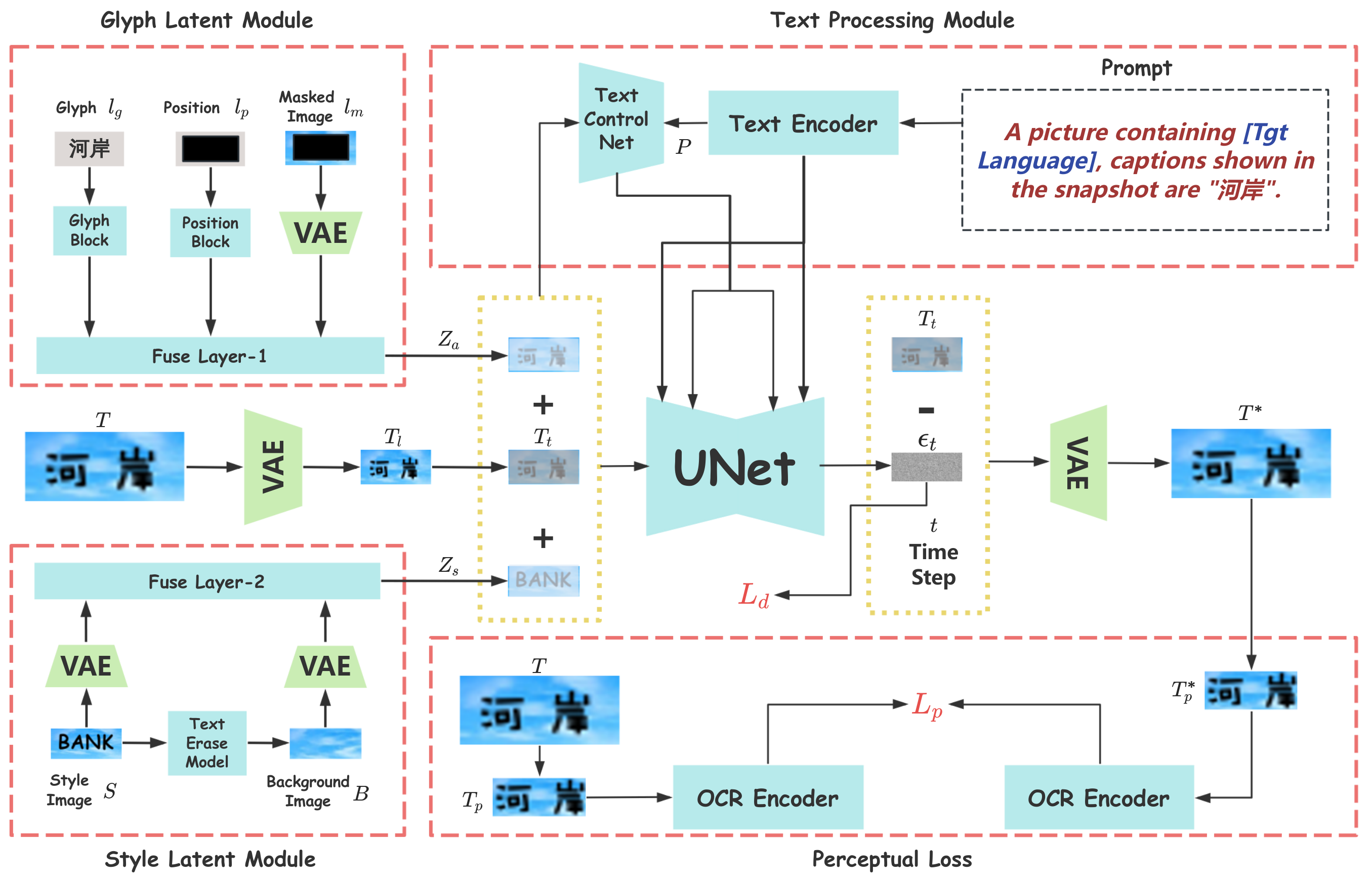} 
\caption{An overview of stage 2. We incorporated a style latent module as a constraint for the Text ControlNet.} 
\label{fig:3} 
\end{figure*}

In the second stage, we employ a diffusion model with text generation enhancement to complete the translation results as shown in figure \ref{fig:3}. The current state of diffusion models can generate high-quality images based on descriptions, but their performance in images with text generation is not satisfactory. Some research has focused on improving the quality of the image with text generation, but in the context of IIT, these efforts still do not fully meet the requirements, which include maintaining style consistency and background fluency. Therefore, we propose a diffusion model that can generate images with consistent text style while preserving the background as much as possible.

\paragraph{Style Latent Module}
If we have pairs of stylistically consistent images, we can input the target image $T$ along with its corresponding style image $S$ as conditions into the text-controlled diffusion process, enabling the model to reference the style of the text in the style image while generating the image. However, in real-life scenarios, it is rare to have such pairs of images that are both stylistically and background-consistent, making the collection of genuine images a costly endeavor. Therefore, we resort to fabricating pairs of style-consistent images to facilitate training, a method we will elaborate on in the experimental section. Additionally, to maintain background consistency as much as possible, we eliminate text from the style image $S$ to obtain pure background information $B$, which is then fed into the model. We employ a VAE decoder $D$ for downsampling both the style image $S$ and the background image $B$, followed by a convolutional fusion layer $g$ to merge these representations and obtain $Z_s$:
$$
Z_s = g(D(S)+D(B))
$$

\paragraph{Glyph Latent Module} 


Consistent with Anytext, we also adopt three auxiliary conditions: glyph $l_g$, position $l_p$, and masked image $l_m$ and employ several stacked convolutional layers to construct mapping layers $G$ and $P$ for downsampling glyph $l_g$ and position $l_p$, utilize a VAE decoder $D$ for downsampling masked image $l_m$, and then integrate this information using a convolutional fusion layer ${f}$ to obtain $Z_a$:

$$
Z_a = f(G(l_g)+P(l_p)+D(l_m))
$$

\paragraph{Text-control Diffusion Pipeline} 
Upon completing the preparation of the style-background information auxiliary module and the text generation auxiliary module, we integrate them as conditions into the text-controlled diffusion pipeline. For an image $T\in\mathbb{R}^{H\times W\times 3}$, we encode it using a Variational Autoencoder(VAE) to obtain latent space feature $T_l\in\mathbb{R}^{h\times w\times c}$. Consistent with the diffusion model, we incrementally add randomly generated Gaussian noise $\epsilon$ to $T_l$ at each time step $t$, creating a noisy latent image $T_t$. Using the representations of the style-background auxiliary module $Z_s$ and the text generation auxiliary module $Z_a$ as conditions, along with the prompt $P$ and time step $t$, the text control diffusion model uses a network $\epsilon_\theta$ to predict noise. The text-control diffusion loss is:
$$
\mathcal{L}_{d} = \mathbb{E}_{T_l,Z_s,Z_a,P,t\sim\mathcal{N}(0,1)}[\lVert\epsilon-\epsilon_\theta(T_t,Z_s,Z_a,P,t)\rVert_2^2]
$$

We also use a text perceptual loss $\mathcal{L}_p$ used in Anytext to further improve the accuracy of text generation. By incorporating the $\mathcal{L}_p$, the model can place greater emphasis on the accurate generation of characters. The final loss $\mathcal{L}$ is expressed as:
$$
\mathcal{L} = \mathcal{L}_d +\lambda * \mathcal{L}_p
$$
$\lambda$ is used to adjust the weight between two loss functions.

\section{Experiments}
In this section, we will progressively introduce the experimental dataset, settings, baselines, and the results.

\subsection{Datasets}
In the second stage, we need to construct pseudo image pairs to complete the experiment. Building upon text synthesis technology, we create stylistically consistent cross-lingual text. To ensure style diversity, we incorporate more than 20 font styles and randomly select text colors, sizes, deformations, and other attributes during text generation. For backgrounds, we randomly crop images to use as backgrounds. As for text content, to resemble real images more closely, we choose parallel corpora from the dataset of the TIT task. We conduct experiments on English to French (En-Fr) and English to Chinese (En-Zh) directions. We generate 100,000 pairs of parallel images for training and 200 pairs for testing in each of the four directions.

To test the effectiveness of our first stage in integrating image information during translation, we have chosen the CoMMuTE \cite{futeral2023tackling} dataset for evaluation. This dataset is specifically designed to address ambiguity in text translation by incorporating images. We also extract a few real images from the OCRMT30K dataset and generate a few images utilizing by text-to-image tool for testing purposes.

\subsection{Baselines and Settings}
We compared our method with existing online systems and Anytrans. Anytrans is a two-stage framework for text translation on images that does not require training. It utilizes MMLLM for translation and employs AnyTrans for backfilling. For online systems, we selected industry-leading Google Image Translate(GoogleTrans), Alibaba Cloud Image Translate(AliyunTrans) and Youdao Image Translation(YoudaoTrans) for comparison.


We utilize Qwen-VL-Chat model, a MMLLM known for its strong capabilities in many languages, as well as its robust multimodal abilities. All experiments are conducted on 4 V100 GPUs. In the second stage, our experimental setup involved a learning rate of 2e-5, a batch size of 6, and training for 15 epochs. We set the parameter $\lambda$ in the loss function to 0.01, consistent with Anytext.

\begin{table}[]
\centering
\begin{tabular}{@{}c|cc|cc@{}}
\toprule
\multirow{2}{*}{\textbf{Method}} & \multicolumn{2}{c|}{\textbf{BLEU}} & \multicolumn{2}{c}{\textbf{COMET}} \\ \cmidrule(l){2-5} 
 & \textbf{Original} & \textbf{Our Method} & \textbf{Original} & \textbf{Our Method} \\ \midrule
\textbf{En-De} & 16.04 & \textbf{17.31} & 72.56 & \textbf{73.84} \\ \cmidrule(r){1-1}
\textbf{En-Fr} & 19.06 & \textbf{20.18} & 72.80 & \textbf{75.23} \\ \cmidrule(r){1-1}
\textbf{En-Zh} & 18.88 & \textbf{19.65} & 80.51 & \textbf{81.56} \\ \bottomrule
\end{tabular}
\caption{Translation results on En-Fr, En-De, En-Zh directions of CoMMuTE Datasets.}
\label{tab:table1}
\end{table}

\begin{figure}[] 
\centering 
\includegraphics[width=0.5\textwidth]{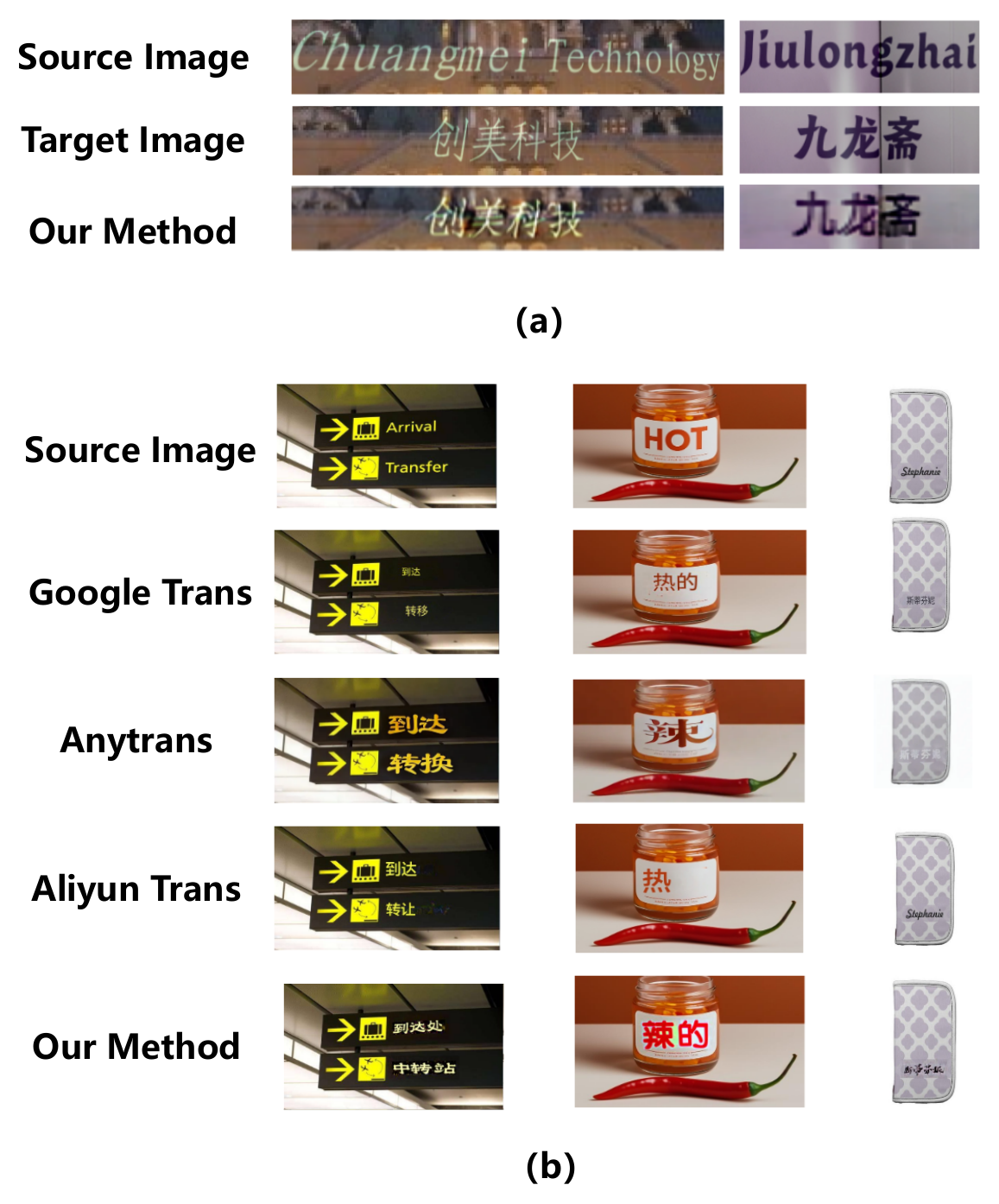} 
\caption{Case study on En-Zh in-image translation on (a) constructed images and (b) real images with our method, AnyTrans, GoogleTrans and AliyunTrans.} 
\label{fig:5} 
\end{figure}

For evaluation metrics, we use the BLEU \cite{papineni2002bleu} and the COMET \cite{rei2020comet} to evaluate the results of machine translation. On the constructed test set with target images, we utilize image similarity metrics such as SSIM \cite{wang2019spatial} and L1 distance. We also employ evaluation methods involving multimodal large models and human assessments. 

\begin{table}[t]
\centering
\begin{tabular}{@{}c|c|cccc@{}}
\toprule
\textbf{} & \textbf{Language} & \multicolumn{1}{c|}{\textbf{\begin{tabular}[c]{@{}c@{}}Aliyun\\ Trans\end{tabular}}} & \multicolumn{1}{c|}{\textbf{\begin{tabular}[c]{@{}c@{}}Youdao\\ Trans\end{tabular}}} & \multicolumn{1}{c|}{\textbf{Anytrans}} & \textbf{HCIIT} \\ \midrule
\multirow{4}{*}{\textbf{SSIM}} & \textbf{En-Zh} & 0.605 & 0.516 & 0.381 & \textbf{0.744} \\
 & \textbf{Zh-En} & 0.657 & 0.622 & 0.439 & \textbf{0.685} \\
 & \textbf{En-Fr} & 0.581 & 0.486 & 0.356 & \textbf{0.668} \\
 & \textbf{Fr-En} & - & 0.467 & 0.281 & \textbf{0.675} \\ \midrule
\multirow{4}{*}{\textbf{L1}} & \textbf{En-Zh} & 0.436 & 0.561 & 0.561 & \textbf{0.402} \\
 & \textbf{Zh-En} & \textbf{0.363} & 0.418 & 0.418 & 0.399 \\
 & \textbf{En-Fr} & 0.476 & 0.588 & 0.588 & \textbf{0.445} \\
 & \textbf{Fr-En} & - & 0.625 & 0.625 & \textbf{0.467} \\ \bottomrule
\end{tabular}
\caption{Our approach, Anytrans, and several online systems' results on the synthesis parallel corpus dataset. AliyunTrans don't support Fr-En Translation Direction.}
\label{tab:table2}
\end{table}

\subsection{Results and Analysis}

\paragraph{Translation Results} 
We present the results of TIT in Table \ref{tab:table1} and juxtapose them with the initial phase of Anytrans (the most effective method for the current TIT task.
). Anytrans predominantly employs a few-shot learning approach ('Original' in Table \ref{tab:table1}), utilizing a large-scale language model to accomplish translation, potentially overlooking the significance of image data. In contrast, our methodology accentuates the amalgamation of image data during the translation process, obviating the need for additional training to seamlessly incorporate image information. Our experimentation conducted on the CoMMuTE dataset, utilizing the Qwen-VL-Chat model as the foundational model, demonstrates superior translation efficacy. The results underscore our method's ability to leverage image data to resolve ambiguities effectively, thereby validating the efficacy of our approach.

\paragraph{Results on Synthesis Test Sets}
To assess the style consistency of generated images, we construct a pseudo test set for evaluation. We conduct tests in four directions across two language pairs: English-French, English-Chinese. As shown in Table 2, we test the effectiveness of our method and existing methods on this dataset, comparing it with online systems such as AliyunTrans, YoudaoTrans\footnote{We don't compare our method with GoogleTrans as they do not offer an official image translation API. GoogleTrans also falls short in terms of consistency, as demonstrated in the case study.}, and recent methods like Anytrans. Within the fabricated dataset, due to the presence of our curated reference images, we can gauge the textual style consistency to some extent by evaluating the image similarity between the generated images and the designated reference images. As shown in Table \ref{tab:table2}, it is observed that the performance of Anytrans is subpar, possibly due to Anytrans predominantly relying on Anytext, which may not perform well when generating text images with large text regions in our synthesis datasets. Our model is trained specifically on this type of data, resulting in superior performance. In comparison to online systems, our method also yield favorable results. This is because these systems do not consider style consistency during translation, leading to lower image similarity. Some examples can be find in figure \ref{fig:5}(a).


\paragraph{Case Study} 
We also apply our method to real datasets. As shown in Figure \ref{fig:5}(b), we present the results of En-Zh translation, comparing them with Google Translate, Anytrans, and Aliyun Translate. In terms of translation consistency, it was observed that other methods exhibit some discrepancies. For instance, translating "Hot" in example as \begin{CJK*}{UTF8}{gbsn} "热的" \end{CJK*}(scorching) instead of the correct \begin{CJK*}{UTF8}{gbsn} "辣的" \end{CJK*}(spicy). Anytrans shows improvements compared to online systems as it uses text features during translation, but doesn't emphasize the integration of image information. In contrast, our method demonstrates stronger integration of image context during translation.

\begin{figure}[t] 
\centering 
\includegraphics[width=0.5\textwidth]{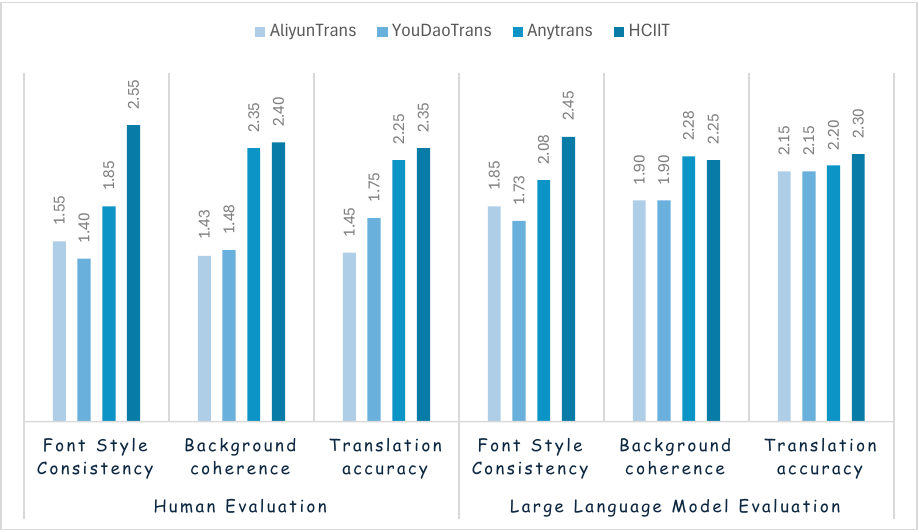} 
\caption{Human evaluation and large language model evaluation.} 
\label{fig:6} 
\end{figure}

Regarding text image generation consistency, it is apparent that commercial online systems tend to remove the original text from the image and render the new text in a standard font, leading to incoherence between the text and the background. AliyunTrans sometimes doesn't translate the text, possibly due to failed OCR. Although AnyTrans generally maintains relatively consistent backgrounds, its performance in text style consistency can sometimes be poor. This is because it fills text based on image background information, which can sometimes introduce additional elements. Our method demonstrates superior performance in text-image style consistency while sometimes the coherence of the background may be compromised.
\paragraph{Human Evaluation and Large Language Model Evaluation}
We also randomly select 40 images from real and fabricated datasets for manual assessment and evaluation using GPT-4o. Our evaluation metrics comprised three components: translation accuracy, background coherence, and font style consistency. Each metric have three levels, with higher scores indicating better performance in that aspect. As depicted in Figure \ref{fig:6}, the diffusion-based methods exhibit good background coherence, yet it still encounters some instances of generation failures. On the other hand, commercial systems demonstrate weak background coherence but maintain a level of stability. Both manual and large model evaluations consistently demonstrate favorable results for our method, showcasing higher translation accuracy and relatively better style consistency.

\section{Conclusion and Future Work}
In this work, we highlight the existing deficiencies in addressing consistency within in-image translation tasks, specifically focusing on translation consistency and image generation consistency. To enhance these aspects of consistency, we propose a two-stage framework comprising text image translation based on MMLLM and image backfilling based on diffusion model. In the first stage, we incorporate CoT learning to improve translation consistency, while in the second stage, we curate a dataset of over 400,000 parallel corpora pairs to train a diffusion model capable of generating text images with consistent styles. This approach enhances the style consistency and background coherence when generating image texts. Ultimately, our framework yields promising results on both constructed and real image test sets, underscoring its efficacy. In the future, we will further progress in image translation from the following perspectives: one is enhancing the layout and other details of text within images, while the other involves researching end-to-end methods.

\bibliographystyle{IEEEbib}
\bibliography{icme2025references}

\begin{thebibliography}{}

\end{thebibliography}


\begin{thebibliography}{10}

\bibitem{ma2022improving}
Cong Ma, Yaping Zhang, Mei Tu, Xu~Han, Linghui Wu, Yang Zhao, and Yu~Zhou,
\newblock ``Improving end-to-end text image translation from the auxiliary text translation task,''
\newblock in {\em 2022 26th International Conference on Pattern Recognition (ICPR)}. IEEE, 2022, pp. 1664--1670.

\bibitem{lan2023exploring}
Zhibin Lan, Jiawei Yu, Xiang Li, Wen Zhang, Jian Luan, Bin Wang, Degen Huang, and Jinsong Su,
\newblock ``Exploring better text image translation with multimodal codebook,''
\newblock in {\em Proceedings of the 61st Annual Meeting of the Association for Computational Linguistics (Volume 1: Long Papers)}, 2023, pp. 3479--3491.

\bibitem{qian-etal-2024-anytrans}
Zhipeng Qian, Pei Zhang, Baosong Yang, Kai Fan, Yiwei Ma, Derek~F. Wong, Xiaoshuai Sun, and Rongrong Ji,
\newblock ``{A}ny{T}rans: Translate {A}ny{T}ext in the image with large scale models,''
\newblock in {\em Findings of the Association for Computational Linguistics: EMNLP 2024}, Nov. 2024, pp. 2432--2444.

\bibitem{caglayan2016multimodal}
Ozan Caglayan, Lo{\"\i}c Barrault, and Fethi Bougares,
\newblock ``Multimodal attention for neural machine translation,''
\newblock {\em arXiv preprint arXiv:1609.03976}, 2016.

\bibitem{yao2020multimodal}
Shaowei Yao and Xiaojun Wan,
\newblock ``Multimodal transformer for multimodal machine translation,''
\newblock in {\em Proceedings of the 58th annual meeting of the association for computational linguistics}, 2020, pp. 4346--4350.

\bibitem{yin2020novel}
Yongjing Yin, Fandong Meng, Jinsong Su, Chulun Zhou, Zhengyuan Yang, Jie Zhou, and Jiebo Luo,
\newblock ``A novel graph-based multi-modal fusion encoder for neural machine translation,''
\newblock in {\em Proceedings of the 58th Annual Meeting of the Association for Computational Linguistics}, 2020, pp. 3025--3035.

\bibitem{caglayan2021cross}
Ozan Caglayan, Menekse Kuyu, Mustafa~Sercan Amac, Pranava~Swaroop Madhyastha, Erkut Erdem, Aykut Erdem, and Lucia Specia,
\newblock ``Cross-lingual visual pre-training for multimodal machine translation,''
\newblock in {\em Proceedings of the 16th Conference of the European Chapter of the Association for Computational Linguistics: Main Volume}, 2021, pp. 1317--1324.

\bibitem{li2022vision}
Bei Li, Chuanhao Lv, Zefan Zhou, Tao Zhou, Tong Xiao, Anxiang Ma, and Jingbo Zhu,
\newblock ``On vision features in multimodal machine translation,''
\newblock in {\em Proceedings of the 60th Annual Meeting of the Association for Computational Linguistics (Volume 1: Long Papers)}, 2022, pp. 6327--6337.

\bibitem{watanabe1998translation}
Yasuhiko Watanabe, Yoshihiro Okada, Yeun-Bae Kim, and Tetsuya Takeda,
\newblock ``Translation camera,''
\newblock in {\em Proceedings. Fourteenth International Conference on Pattern Recognition (Cat. No. 98EX170)}. IEEE, 1998, vol.~1, pp. 613--617.

\bibitem{chen2015integrating}
Jinying Chen, Huaigu Cao, and Premkumar Natarajan,
\newblock ``Integrating natural language processing with image document analysis: what we learned from two real-world applications,''
\newblock {\em International Journal on Document Analysis and Recognition (IJDAR)}, vol. 18, pp. 235--247, 2015.

\bibitem{lan2024translatotron}
Zhibin Lan, Liqiang Niu, Fandong Meng, Jie Zhou, Min Zhang, and Jinsong Su,
\newblock ``Translatotron-v (ison): An end-to-end model for in-image machine translation,''
\newblock {\em arXiv preprint arXiv:2407.02894}, 2024.

\bibitem{ma2023glyphdraw}
Jian Ma, Mingjun Zhao, Chen Chen, Ruichen Wang, Di~Niu, Haonan Lu, and Xiaodong Lin,
\newblock ``Glyphdraw: Learning to draw chinese characters in image synthesis models coherently,''
\newblock {\em arXiv preprint arXiv:2303.17870}, vol. 2, 2023.

\bibitem{yang2024glyphcontrol}
Yukang Yang, Dongnan Gui, Yuhui Yuan, Weicong Liang, Haisong Ding, Han Hu, and Kai Chen,
\newblock ``Glyphcontrol: Glyph conditional control for visual text generation,''
\newblock {\em Advances in Neural Information Processing Systems}, vol. 36, 2024.

\bibitem{chen2024textdiffuser}
Jingye Chen, Yupan Huang, Tengchao Lv, Lei Cui, Qifeng Chen, and Furu Wei,
\newblock ``Textdiffuser: Diffusion models as text painters,''
\newblock {\em Advances in Neural Information Processing Systems}, vol. 36, 2024.

\bibitem{chen2023textdiffuser}
Jingye Chen, Yupan Huang, Tengchao Lv, Lei Cui, Qifeng Chen, and Furu Wei,
\newblock ``Textdiffuser-2: Unleashing the power of language models for text rendering,''
\newblock {\em arXiv preprint arXiv:2311.16465}, 2023.

\bibitem{tuoanytext}
Yuxiang Tuo, Wangmeng Xiang, Jun-Yan He, Yifeng Geng, and Xuansong Xie,
\newblock ``Anytext: Multilingual visual text generation and editing,''
\newblock in {\em The Twelfth International Conference on Learning Representations}, 2023.

\bibitem{huang2024aligning}
Yichong Huang, Xiaocheng Feng, Baohang Li, Chengpeng Fu, Wenshuai Huo, Ting Liu, and Bing Qin,
\newblock ``Aligning translation-specific understanding to general understanding in large language models,''
\newblock {\em arXiv preprint arXiv:2401.05072}, 2024.

\bibitem{futeral2023tackling}
Matthieu Futeral, Cordelia Schmid, Ivan Laptev, Beno{\^\i}t Sagot, and Rachel Bawden,
\newblock ``Tackling ambiguity with images: Improved multimodal machine translation and contrastive evaluation,''
\newblock in {\em Proceedings of the 61st Annual Meeting of the Association for Computational Linguistics (Volume 1: Long Papers)}, 2023, pp. 5394--5413.

\bibitem{papineni2002bleu}
Kishore Papineni, Salim Roukos, Todd Ward, and Wei-Jing Zhu,
\newblock ``Bleu: a method for automatic evaluation of machine translation,''
\newblock in {\em Proceedings of the 40th annual meeting of the Association for Computational Linguistics}, 2002, pp. 311--318.

\bibitem{rei2020comet}
Ricardo Rei, Craig Stewart, Ana~C Farinha, and Alon Lavie,
\newblock ``Comet: A neural framework for mt evaluation,''
\newblock in {\em Proceedings of the 2020 Conference on Empirical Methods in Natural Language Processing (EMNLP)}, 2020, pp. 2685--2702.

\bibitem{wang2019spatial}
Tianyu Wang, Xin Yang, Ke~Xu, Shaozhe Chen, Qiang Zhang, and Rynson~WH Lau,
\newblock ``Spatial attentive single-image deraining with a high quality real rain dataset,''
\newblock in {\em Proceedings of the IEEE/CVF conference on computer vision and pattern recognition}, 2019, pp. 12270--12279.

\end{thebibliography}

\section{Appendix}
\subsection{Pseudo Parallel Image Pairs.}
\begin{figure}[h] 
\centering 
\includegraphics[width=0.5\textwidth]{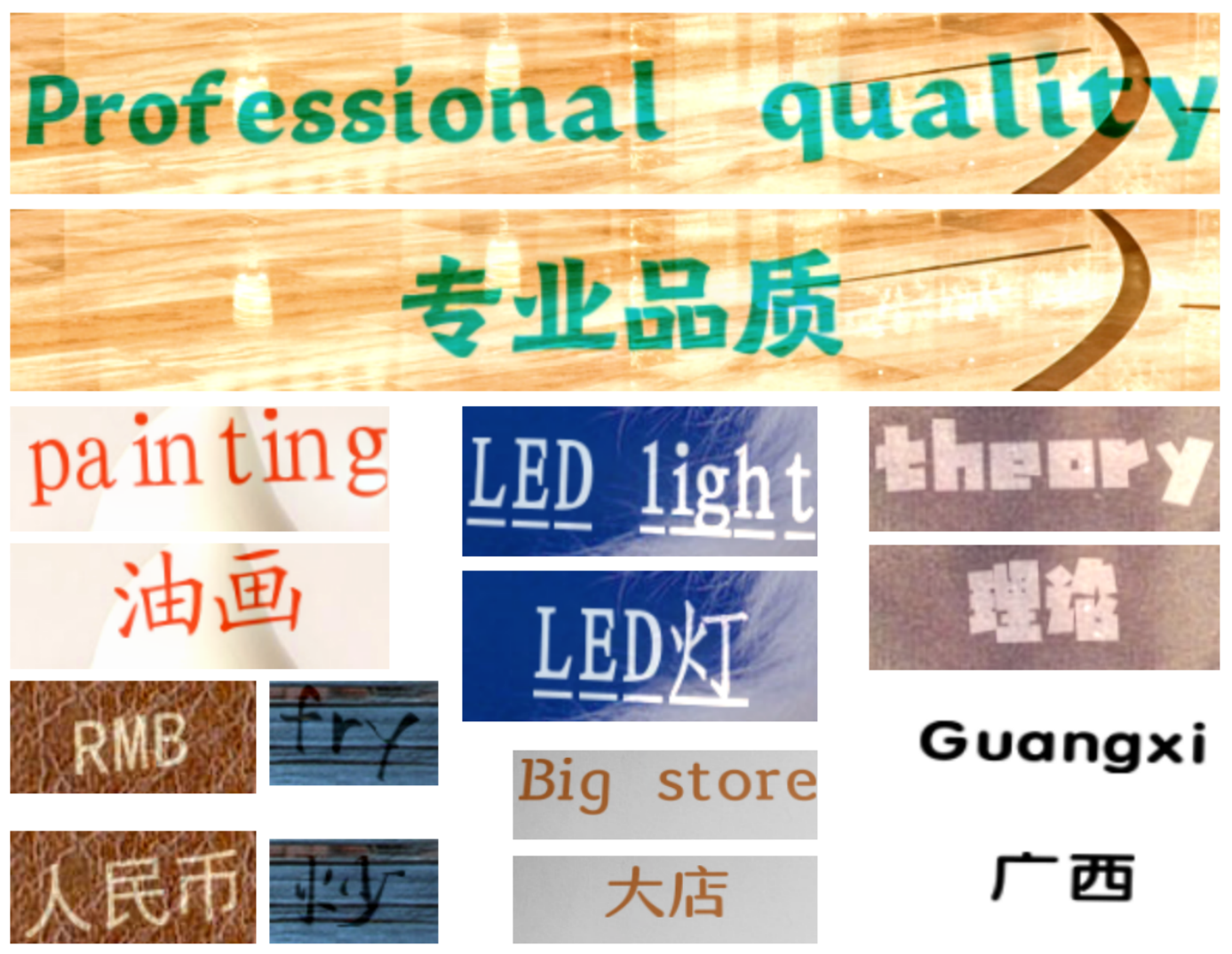} 
\caption{Examples of the constructed pseudo parallel En-Zh image pairs.} 
\label{fig:4} 
\end{figure}

Some examples of the constructed pseudo parallel En-Zh image pairs are shown in Figure \ref{fig:4}.

\subsection{Case Study on Stage 1.}
\begin{figure*}[] 
\centering 
\includegraphics[width=1.0\textwidth]{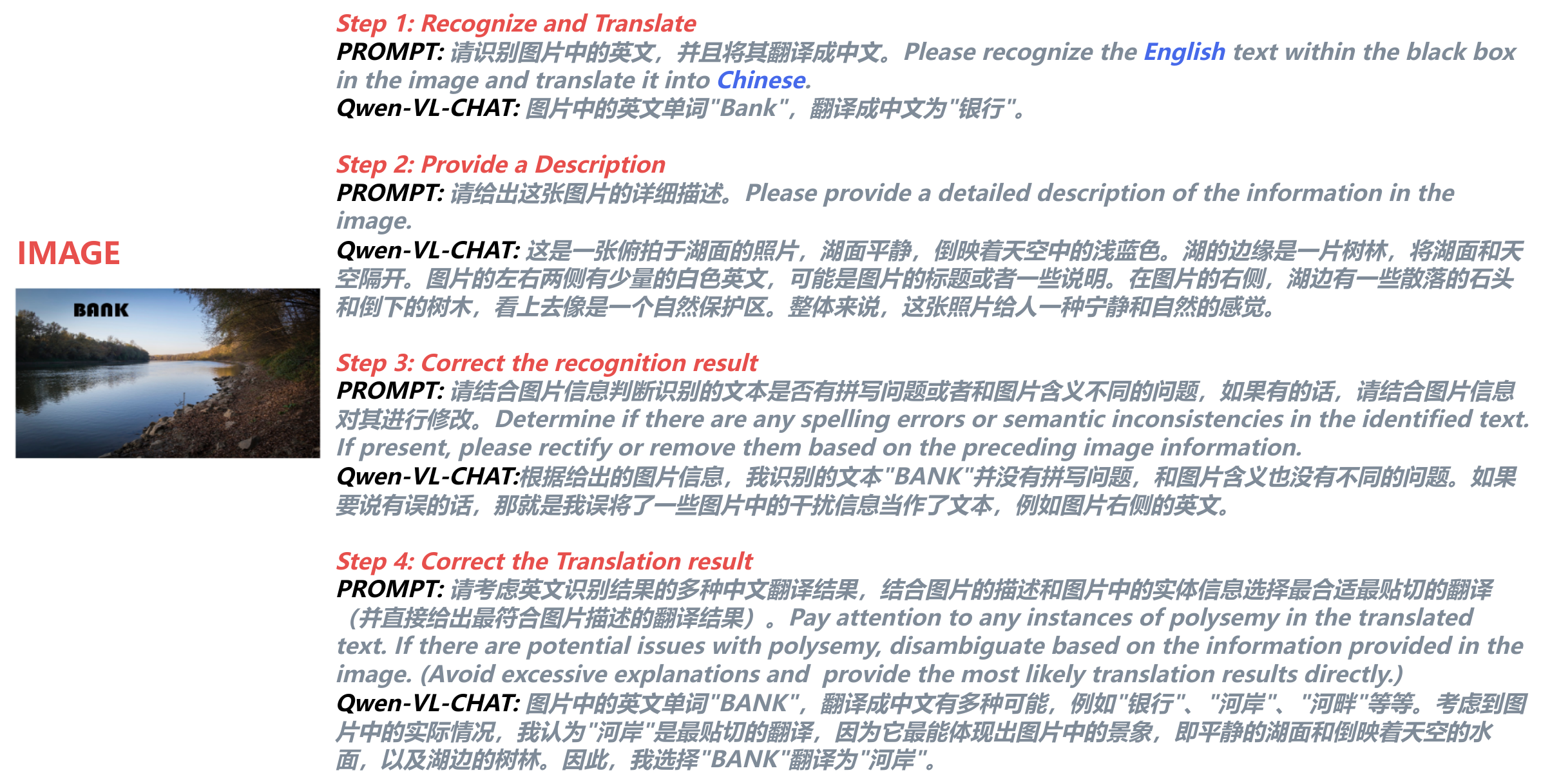} 
\caption{Case Study on Stage 1.} 
\label{fig:7} 
\end{figure*}

We employ the Qwen-VL-Chat model as a multimodal multilingual large-scale model in the first stage. In this section, we present a case study, as depicted in Figure \ref{fig:7}, where the model accurately integrates image information during translation when utilizing our proposed COT method. In contrast, without its application, the model tends to produce incorrect translation results (step 1 in the image). Through experimentation, we have found that utilizing the target language (Chinese here) in our prompts yields superior results compared to using the source language.

\subsection{Text Erase Model.}
As for Text Erase Model, as we employ artificially constructed images, we inherently possess backgrounds devoid of text from the outset. Thus, during training, we directly utilize preexisting background images. In real image testing scenarios, we employ AnyText to carry out the text removal process. In cases where the removal efficacy is subpar, we resort to employing image editing software for further refinement.

\subsection{Human evaluation details}
In our work, the Human evaluation details have drawn inspiration from some of the evaluation specifics used in AnyTrans. Before annotation, we also provide relevant examples to ensure that the annotators understand the annotation details. We have proposed three directions of evaluation metrics for two types of consistencies, each with 3 levels of scores:

1.Translation Accuracy:
\begin{itemize}

   \item - Point 1: Translation errors or failure to recognize text, significant translation inaccuracies, completely irrelevant.
   \item - Point 2: Correct translation but text is blurry or contains errors.
   \item - Point 3: Translation is completely accurate, effectively integrates image information, text are displayed correctly without any issues.
\end{itemize}

2.Font Style Consistency:
\begin{itemize}

   \item - Point 1:Low text style consistency. Text style is completely inconsistent, with variations in color, font, thickness, etc.
   \item - Point 2: Medium text style consistency. Partial consistency in text style, some elements like color, font, thickness are consistent.
   \item - Point 3: High text style consistency. Text style is nearly entirely consistent, color, font, thickness match perfectly, and relevant text details are relatively consistent.
    
\end{itemize}

3.Background Coherence:
\begin{itemize}
   \item - Point 1: Low coherence, clear boundaries between background and text (common in commercial systems), significant sense of layering.
   \item - Point 2: Medium coherence, text can blend into the background without distinct boundaries.
   \item - Point 3: High coherence, text and background seamlessly integrate, background is clear with no sense of boundaries.
\end{itemize}

\subsection{Future Work.}
In the future, we will continue to advance in-image translation from the following perspectives to further enhance its capabilities:
\begin{itemize}

\item Improving the detailed content of IIMT: IIMT is a complex task encompassing multiple modalities and diverse scenarios. While our method has proposed a general framework, there are still intricacies that require resolution within this framework. These include translating vertical text images, translating multi-line text images, collaborative translation of multiple texts, translation of lengthy text images, and optimizing text positioning during translation. Our forthcoming research endeavors will concentrate on refining these aspects.
\item End-to-end IIMT frameworks: Presently, models that exhibit superior performance are often designed in a cascaded manner. Cascaded approaches may encounter issues such as error accumulation and high latency. Therefore, our upcoming research will investigate end-to-end methods for IIMT to overcome these challenges and improve efficiency.

\end{itemize}


\end{document}